# Data-Driven Traffic Assignment: A Novel Approach for Learning Traffic Flow Patterns Using a Graph Convolutional Neural Network

Rezaur Rahman[a] and Samiul Hasan[b]

[a] Department of Civil, Environmental Engineering, and Construction Engineering
University of Central Florida
Email: rezaur.rahman@knights.ucf.edu

[b] Department of Civil, Environmental Engineering, and Construction Engineering
University of Central Florida
Email: samiul.hasan@ucf.edu

**ABSTRACT**

We present a novel data-driven approach of learning traffic flow patterns of a transportation network given that many instances of origin to destination (OD) travel demand and link flows of the network are available. Instead of estimating traffic flow patterns assuming certain user behavior (e.g., user equilibrium or system optimal), here we explore the idea of learning those flow patterns directly from the data. To implement this idea, we have formulated the traffic-assignment problem as a data-driven learning problem and developed a neural network-based framework known as Graph Convolutional Neural Network (GCNN) to solve it. The proposed framework represents the transportation network and OD demand in an efficient way and utilizes the diffusion process of multiple OD demands from nodes to links. We validate the solutions of the model against analytical solutions generated from running static user equilibrium-based traffic assignments over Sioux Falls and East Massachusetts networks. The validation result shows that the implemented GCNN model can learn the flow patterns very well with less than 2% mean absolute difference between the actual and estimated link flows for both networks under varying congested conditions. When the training of the model is complete, it can instantly determine the traffic flows of a large-scale network. Hence this approach can overcome the challenges of deploying traffic assignment models over large-scale networks and open new directions of research in data-driven network modeling.

## 1. INTRODUCTION

Traffic assignment problem (TAP) is one of the key components of transportation planning and operations. It is used to determine traffic flows of each link of a transportation network for a given travel demand based on modeling the interactions among traveler route choices and the congestion that results from their travel over the network (Sheffi, 1985). Traditionally traffic assignment problems have been formulated as mathematical programs and solved based on user equilibrium (UE) principles (Wardrop, 1952). The UE solution relies on several assumptions on user behavior and knowledge such as: (i) drivers have perfect information and knowledge about the underlying network; (ii) drivers make rational choices when choosing a route; and (iii) all drivers are homogeneous (Kim et al., 2009). Although some of these assumptions may not hold in a real-world scenario, this approach has been providing the most reasonable solutions of the traffic assignment problem (Bar-Gera, 2002; Jafari et al., 2017; LeBlanc et al., 1975; Leurent et al., 2011; Mitradjieva and Lindberg, 2013; Tizghadam and Leon-garcia, 2007). One of the major issues with the static traffic assignment solution is that it assumes constant OD demand whereas traffic demand may significantly change over time, which makes it unsuitable to deploy for real world traffic scenario.

Static traffic assignment problem fails to capture traffic dynamics of a network, which motivated researchers to move towards dynamic traffic assignment methods. Dynamic traffic assignment (DTA) has become a state-of-the-art solution method to capture traffic network dynamics in a more realistic way. Numerous formulation and solution approaches have been introduced ranging from dynamic programming to variational inequality to simulation-based approaches (Ban et al., 2008; Boyles et al., 2006; Friesz et al., 1989; Janson, 1989; Jiang et al., 2011; Z. Li et al., 2018; Liu et al., 2007; Lo and Szeto, 2002; Mahmassani, 2001; Merchant and Nemhauser, 1978; Nie and Zhang, 2010; Peeta and Mahmassani, 1995; Peeta and Ziliaskopoulos, 2001; Primer, 2011; Ran et al., 1993; Ukkusuri et al., 2012; Waller et al., 2013; Watling and Hazelton, 2003). In a smaller model with less complexity, the DTA process works well in converging to a point of equilibrium. As the scale of a model grows in size, complexity, and congestion, the DTA procedure may become more difficult and require more computational time making it less suitable for a real-time deployment. Researchers are exploring to develop efficient solution methods for traffic assignment problem (TAP) which can be deployed for real-time transportation operation purposes.

Optimization of a transportation network largely depends on how accurately we capture the traffic dynamics of the network to forecast traffic in real-time. In particular, active traffic management requires a system that can estimate and forecast traffic in real-time. Over the last decade, traffic sensing technologies and emerging connected vehicles environment with added computational power has created an opportunity towards developing a data-driven approach as an alternative to traditional approaches of solving traffic assignment problems. Roadway sensors and devices provide us information on instantaneous travel time variation for different links indicating the congestion propagation throughout the network. Moreover, OD demand can be extracted from emerging data sources such as mobile phones, GPS observations, and social media posts. As such, we now have real-world O-D demand and traffic flow data available at large-scale over many days. However, one challenge is how to develop a data-driven framework at a network scale to learn the relationship between OD demand and link flows.

Existing data-driven approaches to traffic prediction have limitations such as: (i) they predict only short-term traffic states (speed, flow, travel time) (Billings and Jiann-Shiou, 2006; Deshpande and Bajaj, 2016; Lee, 2009; Wu et al., 2004; Yu et al., 2016); (ii) they work over only one or multiple segments of highways, but not at the scale of a network (Cai et al., 2016; Chien and Kuchipudi, 2003; Elhenawy et al., 2014; Innamaa, 2005; Luo et al., 2019; Ma et al., 2015; Myung et al., 2011; Polson and Sokolov, 2017; Rahman and Hasan, 2018; Zhang and Haghani, 2015); and (iii) they do not consider the travel demand variation while predicting future traffic. As such, these approaches consider traffic prediction as a simple time series problem and predict the traffic state for a shorter time horizon (e.g., next 5 to 15 min). However, network level traffic prediction is more challenging due to the higher computational complexity required by the network topology. Such a problem requires (i) to capture the travel demand variation in the network, (ii) to capture the correlation of traffic in interconnected roads and mapping it in a spatial network (Polson and Sokolov, 2017), and (iii) to reflect drivers' route choice and associated congestion propagation inside the network.

Neural network approaches such as Feed Forward and Long Short Term Memory neural networks have been used in numerous short-term traffic state prediction problems, such as, traffic speed prediction (Cui and Wang, 2017; Epelbaum et al., 2017; Ma et al., 2015), travel time prediction (Yanjie Duan et al., 2016), traffic flow prediction (Luo et al., 2019; Polson and Sokolov, 2016; Yang et al., 2019), vehicular queue length prediction (Lee et al., 2019; Rahman and Hasan, 2020). Moreover, Convolutional LSTM neural networks (ConvLSTM) have become as the state-of-the-art approach for network-level traffic forecasting. However, in a ConvLSTM model, the traffic network is considered as an image and network information is extracted using a convolution filter, which is then fed into an LSTM model to predict future traffic state. While traffic network is converted into an image, a certain amount of noise is included into the spatial relationships of traffic state causing erroneous results in prediction. Moreover, ConvLSTM cannot capture the long-term congestion propagation inside the network for long term traffic forecasting.

Recently, graph theory coupled with generalized neural network architecture has been utilized to model the dynamics between the structural properties and the functions of a network (Atwood and Towsley, 2016; Cui et al., 2018; Li et al., 2017; Zhou et al., 2018) solving problems such as modeling physical systems, learning molecular fingerprints, predicting protein interface, and classifying diseases, which require that a model learns from graph-based inputs (Zhou et al., 2018). However, the application of such a neural network architecture hardly exists for a large-scale transportation network. A few studies have proposed as spatial and temporal learning-based methods for small scale traffic prediction (Cui et al., 2018; Li et al., 2017). These methods do not solve the fundamental traffic assignment problem considering flow propagation inside the network. Furthermore, these methods have not been tested in any realistic networks. They have used few interconnected point-based detectors to create a small-scale network and validated the proposed algorithm on that network. Moreover, existing methods do not consider the physical characteristics of roadways such as free flow travel time, length, speed limits, and number of lanes.

In this paper, we present a novel data-driven approach of learning traffic flow patterns of a transportation network given that many instances of OD demand and traffic flow of the network are available. Instead of estimating traffic flow patterns assuming certain user behavior (e.g., user equilibrium), here we explore the idea of learning those patterns from large-scale training data by developing a neural network architecture. In particular, we use the Graph Convolutional Neural Networks (GCNN) which generalize traditional neural networks to work on structured graphs (Kipf and Welling, 2016). GCNNs utilize an adjacency matrix or a Laplacian matrix to represent the structure of a graph.

Recently, several studies have utilized the concept of graph convolution to represent traffic network as a generalized graph for traffic state prediction (Cui et al., 2018; Li et al., 2017). Moreover, Graph Convolution Neural Network has been emerging as a new approach to overcome the challenge of high dimensionality when predicting travel demand for a large-scale network (Kim et al., 2019; Lin et al., 2018). However, this is the first time, we developed a GCNN model to learn the traffic flow pattern of a transportation network through learning the diffusion process of multiple OD demands from nodes to links. In this work, we seek to answer the question: *can a deep learning model learn the flow patterns of a network without relying on the assumptions of user behavior for assigning traffic in the network*? If positive, apart from assigning traffic in a purely data-driven way, this can open new directions in transportation network modeling research offering data-driven approaches for solving dynamic traffic assignment, network design, and many other problems with real-world applications. Thus, this study makes several contributions:

(i) It formulates the traffic-assignment problem as a data-driven learning problem considering the underlying transportation network and available instances of OD demand and link flows, without assuming any user behavior.

(ii) It develops a novel neural network architecture that solves the learning problem to determine link flows based on OD demand and traffic flow data.

(iii) It provides rigorously tested experimental evidence that such a neural network architecture can learn the user equilibrium traffic flow assignment of a transportation network only from data.

## 2. DATA-DRIVEN TRAFFIC ASSIGNMENT

In this section, first we formulate the data-driven traffic assignment problem. We then describe the Graph Convolutional Neural Network (GCNN) approach to model traffic flow propagation in the network.

### 2.1 Problem Definition

In a transportation network, all nodes are connected, and each link is associated with information such as distance, speed limit, capacity etc. Here, we consider the transportation network as a weighted directed graph $\mathcal{G}(v, \mathcal{E}, A_w)$ where $v$ denotes the set of nodes and $\mathcal{E}$ denotes the set of links between nodes $(i,j)$. $A_w$ represents the connectivity between nodes as a weighted adjacency matrix, where weights are based on free flow travel time between any two nodes $(i,j)$, defined as follows:

$$A_w(i,j) = \begin{cases} t_{i,j}^0 & if\ i \to j \\ t_{j,i}^0 & if\ j \to i \\ 0, & if\ i = j \end{cases} \quad (1)$$

where, $t^o$ denotes the free flow travel time between the origin and the destination nodes. The proposed data-driven formulation of the traffic assignment problem aims to *learn* the flow patterns of a transportation network based on the network structure and the instances available on origin to destination (OD) travel demand and link flows (Fig.1). Also, we have the information on network characteristics, such as location of each node with respect to other nodes, travel distance or free flow travel time between different nodes. From this information, we develop a data-driven method to estimate the link flows for given travel demands. During the estimation process, we also learn the traffic flow propagation from origin nodes toward destination nodes in a transportation network. Let, $X$ be the demand matrix for the transportation network $\mathcal{G}$, where each element of a row indicates the travel demand between origin node $(i)$ and the destination node $(j)$.

The traffic assignment problem aims to learn a function $\mathcal{F}(.)$ that maps $m$ instances of OD demand matrix $(X_1, X_2, X_2 \ldots\ldots\ldots, X_m)$ to $m$ instances of flow $(F_1, F_2, F_3 \ldots\ldots\ldots F_m)$, defined as follows,

$$\mathcal{F}([X_1, X_2, X_2 \ldots\ldots\ldots, X_m]; \mathcal{G}(v, \mathcal{E}, A_w)) = [F_1, F_2, F_3 \ldots\ldots\ldots F_m] \quad (2)$$

where, $A_w$ indicates the weighted adjacency matrix, $\mathcal{E}$ indicates the set of links of the network, and the vector $F_m$ contains the link flows for each link of the network for a given OD demand $(X_m)$. In this formulation, OD demands and network properties are input variables and link flows are the target variables.

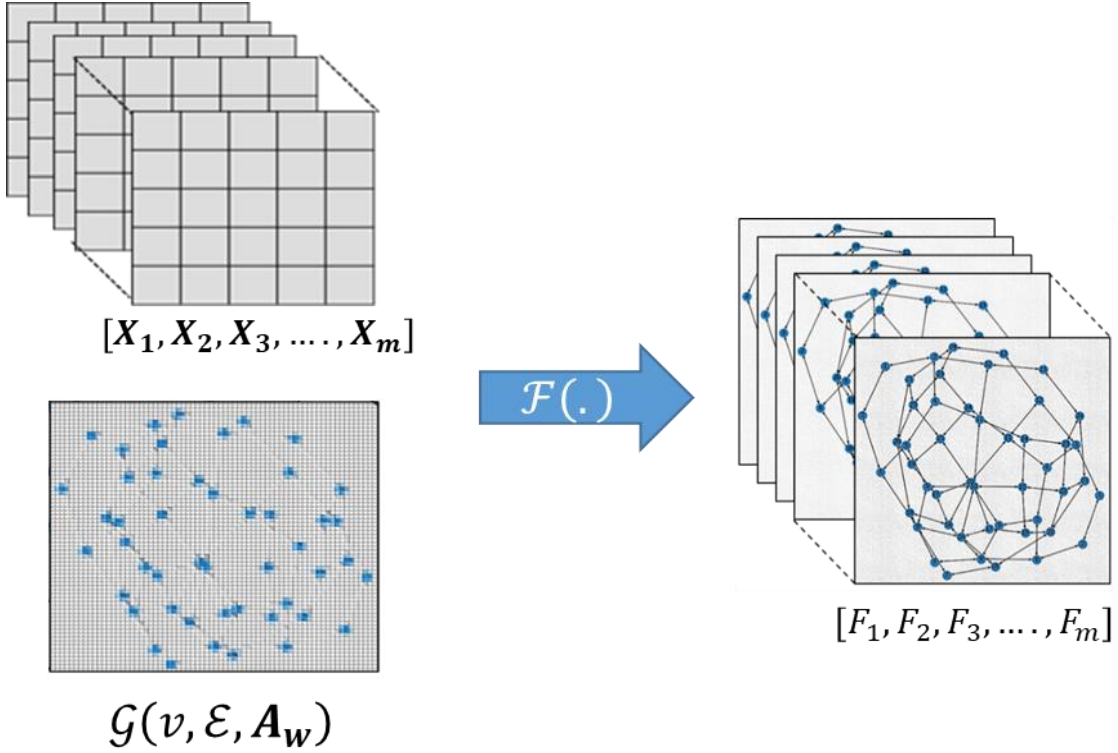

Fig. 1. A Schematic View of the Data-Driven Traffic Assignment Problem

**Table 1** List of notations

| Notation | Description |
|---|---|
| $\mathcal{G}$ | Transportation network |
| $v$ | Set of nodes in $\mathcal{G}$ with size of $|v| = N$ |
| $\mathcal{E}$ | Set of links in $\mathcal{G}$ with size of $|\mathcal{E}| = E$ |
| $A_w \in R^{N \times N}$ | Weighted adjacency matrix of $\mathcal{G}$, defined by Equation (1) |
| $I \in R^{N \times N}$ | Identity matrix |
| $\bar{A}_w \in R^{N \times N}$ | Neighborhood matrix defined by Equation (7) |
| $D_w \in R^{N \times N}$ | Degree matrix of $\mathcal{G}$, a diagonal matrix where diagonal elements $(i,i)$ indicate the number of links coming out from a node |
| $L_w \in R^{N \times N}$ | Laplacian matrix represents the structural properties of a network, defined by the equation (12) |
| $t_{i,j}^0$ | Free flow travel time between nodes $i$ and $j$ |
| $X \in R^{N \times N}$ | OD demand matrix |
| $F \in R^E$ | Flow vector contains flows for each link of the network for a given OD demand $X$ |
| $\mathcal{P} \in R^{N \times N}$ | Routing matrix indicates the probability of diffusion of flow from node $i$ to node $j$ |
| $g_\theta$ | Convolutional filter to learn the network features along with network function |
| $f(.)$ | Activation function |
| $\Theta \in R^{N \times N}$ | Learnable parameters for the convolution filter |
| $W_q, W_F$ | Learnable parameters for link flow estimation |
| $q$ | Flow distribution matrix indicating the propagation of flow from a given node to different links of the network |
| $H$ | Indicates the outputs from different layers of the proposed neural network architecture |

All the bold letters denote a matrix

## 2.2 Graph Convolution Neural Network for Flow Pattern Learning

We develop a Graph Convolution based Neural Network (GCNN) architecture to assign traffic in the transportation network. GCNN generalizes the traditional convolutional neural network approach which uses a random filter (e.g., gaussian) to extract spatial features (e.g., spatial correlations) from the available features of a network. The network features are defined as a matrix, where each element includes specific information about the network (e.g., origin to destination demand). The convolutional operation multiplies a convolution filter with the network feature matrix to capture the cross-correlation among these spatial features. However, the problem with a convolutional neural network is that it considers the transportation network as an image; hence it does not capture how traffic states (e.g., flow) changes inside the network.

     To determine the solution of a data-driven traffic assignment problem, we adopt the concept of graph convolution. We use graph convolution operation to learn the network properties and the flow diffusion process from origin nodes toward destination nodes. To estimate link flows, we model how this flow diffusion process is contributing to link flows. In other words, the model considers how flows are coming to a specific link from adjacent nodes while diffusing from origin nodes towards destination nodes. In the proposed GCNN model, a convolution filter is derived based on the network structure (position of the nodes and links) and the flow diffusion process inside the network. Hence, GCNN model captures changes in traffic states of a transportation network by modeling the flow diffusion from origin nodes toward destination nodes. As such, the GCNN model simultaneously learn the features (each node and link are embedded with valuable information) and function of the network.

     Once the model learns the flow diffusion process, we can feed this information into a feed forward neural network to estimate the traffic flows at different links. In the following section, we describe the final architecture of the proposed deep neural network model to estimate traffic flows at different links.

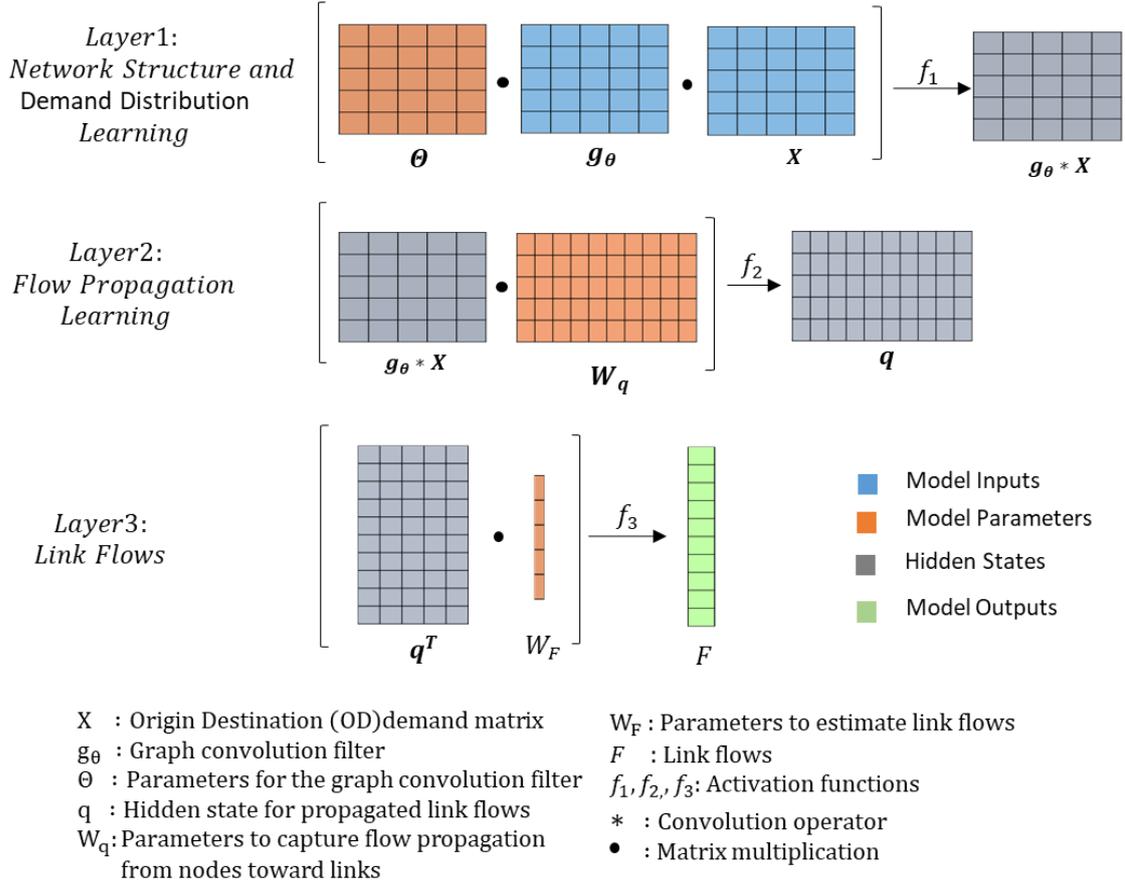

Fig. 2. GCNN architecture for traffic flow pattern learning

Fig. 2 shows all the layers and matrix operations in each layer for the proposed GCNN architecture. In the first layer, we define the graph convolutional operation to learn the network properties and demand distribution for different nodes. We derive the graph convolution filter from weighted adjacency matrix where weights are assigned based on the free flow travel time of a link (see equation 1). We perform the convolution operation between the OD demand ($X$) and graph convolution filter ($g_\theta$). The convolutional filter ($g_\theta$) represents the diffusion process of the traffic flows from the origin node towards destination nodes. While training the model, we estimate the parameters ($\Theta$) for this convolutional filter. We define graph convolution layer as follows,

$$\mathbf{H^1} = f_1(g_\theta * X) = f_1(\Theta g_\theta X) \qquad (3)$$

here, $f_1(.)$ denotes the nonlinear activation function for the convolution layer and $\mathbf{H^1} \in R^{N \times N}$ indicates the output from the graph convolution layer (1st layer). From this layer, we obtain a convoluted demand matrix ($\mathbf{H^1}$) representing the flow diffusion process (from origin nodes towards destination nodes) inside a network.

The convoluted demand matrix is then fed into the 2nd layer of the GCNN, where we model traffic flow distribution from origin nodes towards adjacent links. In this layer, we create a simple neural network model with parameters $W_q$, which maps the convoluted demand matrix to a $N \times E$ dimensional space (same size as the link-node adjacency matrix) via matrix multiplication. In this way, the GCNN captures how flow diffusion process will assign traffic at different links of the network. We define the 2nd layer of the model as follows,

$$\mathbf{H}^2 = \mathbf{q} = f_2(W_q \mathbf{H}^1) \tag{4}$$

here, $f_2(.)$ denotes the nonlinear activation function for the second layer and $\mathbf{H}^2(=\mathbf{q}) \in R^{N \times E}$ indicates the output from the second layer representing distributed link flows from adjacent origin nodes (N) of the network. Which means, each row of the matrix $\mathbf{q}$, indicates the distributed link flows for all the links (E) associated with an origin node.

Finally, the distributed link flow matrix ($\mathbf{q}$) is transposed and fed into the output layer (see Fig. 2). Inside the transposed matrix ($\mathbf{q}^T \in R^{E \times N}$) each row indicates the distributed link flows for a given link from all the origin nodes (N). In the output layer, we assign a linear activation function ($f_3(.)$) with N parameters, which aggregates the distributed link flows and outputs assigned traffic flow for a given link. We define the output layer as follows,

$$\mathbf{H}^3 = \mathrm{F} = f_3(W_F(\mathbf{H}^2)^T) = f_3(W_F \mathbf{q}^T) \tag{5}$$

Here, $f_3$ denotes the linear activation function and $\mathrm{H}^3(=F) \in R^E$ denotes the assigned traffic flows for all the links. From the output layer ($\mathrm{H}^3$), we obtain link flows (F) for a given OD demand (X). So, the mathematical formulation of the GCNN model to estimate the link flows can be generalized as follows,

$$F = f_3((f_2(f_1(\Theta g_\theta X) W_q)^T W_F) \tag{6}$$

here, we select the hyperbolic tangent function $tanh = \frac{e(x)-e(-x)}{e(x)+e(-x)}$ as the nonlinear activation function ($f_1(.) = f_2(.) = tanh$) for the model.

In the following section, we describe the graph convolution operations and the parameters associated with graph convolution filter in details.

## 2.3 Modeling Traffic Flow using Diffusion Graph Convolution

In a transportation network, the traffic flow pattern changes in response to the changes in travel demand. We represent this relationship via flow diffusion process from origin nodes towards destination nodes. To capture the stochastic nature of traffic flow variation at a network level, we consider the flow diffusion process by a random walk (random movement into adjacent neighboring nodes) in the network, $\mathcal{G}$ with restart probability $\alpha \in [0,1]$ and a state transition matrix $\overline{D}_w^{-1}\overline{A}_w$, where $\overline{A}_w$ is the neighborhood matrix. In the neighborhood matrix, we add an identity matrix (I) with the adjacency matrix ($A_w$). By adding the identity matrix, we create a self-loop for each node; for a given diffusion step it will capture the traffic flows having same node as origin and destination; in other words, it captures that the traffic flow remains in the origin node rather than diffusing from origin node to destination nodes. The neighborhood matrix $\overline{A}_w$ is defined as,

$$\overline{A}_w = A_w + I \tag{7}$$

The restart probability indicates the probability of starting of a random walk from node $i$. From the starting node such random walks take multiple steps (diffusion steps, K) to traverse the adjacent nodes until reaching the destination node $j$. After many time steps, such diffusion process converges to a stationary distribution $\mathcal{P} \in R^{N \times N}$, where $i$th row of $\mathcal{P}$ indicates the probability of flow diffusion from node $i$ towards $j$. The stationary distribution of the diffusion process can be represented as a weighted combination of infinite random walks on the graph (Teng, 2016) and be calculated in closed form,

$$\mathcal{P} = \sum_{k=0}^{\infty} \alpha(1-\alpha)^k (\overline{D}_w^{-1}\overline{A}_w)^k \tag{8}$$

where $k$ is the diffusion step. In practice, we can consider a finite number of diffusion steps and assign a trainable weight at each step. Similar diffusion processes have been adopted in previous research (Atwood and Towsley, 2016; Y. Li et al., 2018). Based on that we can define the diffusion convolution over the network features (e.g., OD demand) $X$ as follows,

$$g_\theta * X = \sum_{k=0}^{K-1} \Theta_k (\overline{D}_w^{-1}\overline{A}_w)^k X \tag{9}$$

If we consider a 2-step diffusion process the above equation becomes,

$$g_\theta * X = \Theta_0 X + \Theta_1 (\overline{D}_w^{-1}\overline{A}_w)^1 X \tag{10}$$

where, $\Theta_0$ and $\Theta_1$ are the weights for each of the steps. While modeling the traffic flow pattern of transportation network, we can consider different number of diffusion steps to reach to a stationary distribution. However, as the network grows, this process becomes computationally expensive, since for a larger network the value for the diffusion step will be higher. Hence, such diffusion processes are only applicable for small scale networks, where flow diffusion occurs among the nearest neighbors (Li et al., 2017; Wang et al., 2020).

In our problem, we propose an alternative approach to represent the diffusion process. Instead of selecting a value of the diffusion step ($K$) and assigning parameter to each step, we assign parameters ($\Theta$) to locally learn the stationary probability distributions (probability matrix). So, while the training the model we estimate the parameters ($\Theta$) to learn the stationary probability distribution for the flow diffusion process. In other words, we obtain a routing matrix (Leon-Garcia and Tizghadam, 2009) which indicates the probability of diffusion of flow from node $i$ to node $j$. The resulted diffusion convolution over the OD demand $X$ can be written as follows:

$$g_\theta * X = \Theta(\overline{D}_w^{-1}\overline{A}_w)X \tag{11}$$

where, $\Theta \in R^{N \times N}$ are the parameters of the convolution filter and $\overline{D}_w^{-1}\overline{A}_w$ represents the transition probability matrix of the diffusion process. In the demand matrix $X$, each row indicates travel demand from origin node $i$ to destination node $j$. So, when we perform the matrix multiplication between the operator $\Theta(\overline{D}_w^{-1}\overline{A}_w)$ and $X$, we obtain a convoluted feature matrix which captures the influence all OD pairs on link flows associated with origin node.

We can also model the diffusion process using a normalized Laplacian matrix. Laplacian matrix better represents the structural properties of a network: the diagonal elements indicate the number of links originating at a given node, while the other elements indicate the connection between the origin and destination nodes. We define the Laplacian matrix as follows,

$$\bar{L}_w = \bar{D}_w - \bar{A}_w \qquad (12)$$

Normalizing the Laplacian matrix with degree matrix,

$$\bar{D}_w^{-1}\bar{L}_w = \bar{D}_w^{-1}(\bar{D}_w - \bar{A}_w) = I - \bar{D}_w^{-1}\bar{A}_w \qquad (13)$$

Now, the convolution over OD Demand matrix, $X$ can be written as follows,

$$g_\theta * X = \Theta(I - \bar{D}_w^{-1}\bar{A}_w)X \qquad (14)$$

where, $\Theta \in R^{N \times N}$ are the parameters of the convolution filter, in other words, the coefficient matrix of the diffusion equation. During the training of deep learning model, we learn these parameters, which capture the flow diffusion process inside the network.

In this study, we focus on a probabilistic approach to model the flow diffusion by estimating transition probability matrix in two ways: using a random walk on adjacency matrix (equation 11) and Laplacian matrix (equation 14). We also compare these approaches with spectral graph convolutional neural network (Kipf and Welling, 2016) to learn traffic flow patterns. In a spectral graph convolutional approach, the convolutional filter is estimated by decomposing the adjacency matrix into its eigenvalues to represent different properties of the graph such as strength of a node, shortest path distance etc. In Appendix A, we provide the details of the concepts of spectral graph convolution to learn traffic flow patterns of a transportation network.

## 3. DATA GENERATION

Although we propose this method for real-world traffic data, recent sensing technologies are not densely distributed yet to provide us data necessary to test this approach. Especially, for large networks the OD demand variations are not accessible to us, though such OD demand data exist due to the availability of mobile phone data (Alexander et al., 2015; Gundlegård et al., 2016). In addition, we seek to verify if the proposed approach works across different congested conditions and to measure the gaps between the actual traffic assignment solutions (i.e., analytical solutions) and the solutions obtained from the proposed neural network-based approach. Thus, to verify our approach, we generate synthetic traffic data based on user equilibrium (UE) solutions of static traffic assignments over two networks: Sioux Falls network (24 nodes and 76 links) and East Massachusetts Network (74 nodes and 258 links). It should be noted that our approach is not an alternative method to determine UE solutions. We use the UE based traffic assignment mainly to generate the data to verify our approach.

We obtained the OD demand and information on network characteristics from (Transportation Networks for Research Core Team, 2016). To generate multiple OD demand, we multiplied the OD demand matrix by random factors collected from a uniform distribution which varies 0.1 to 1.0. To test our approach in different scenarios, we consider three conditions: uncongested, moderately congested, and fully congested. We generate 5,000 OD matrices for each condition and solve both networks using the Frank Wolfe algorithm to obtain user equilibrium traffic flows. To represent the prevailing traffic condition, we estimate the flow over capacity ratio. We assume that, for uncongested condition the flow-capacity ratio remains less than 0.5, for moderate condition the flow-capacity ratio varies between 0.4 and 0.8, and for uncongested condition, most of the cases the flow-capacity ratio is greater than 1.0. Fig. 4 shows the traffic flow variations for different links of Sioux Falls Network.

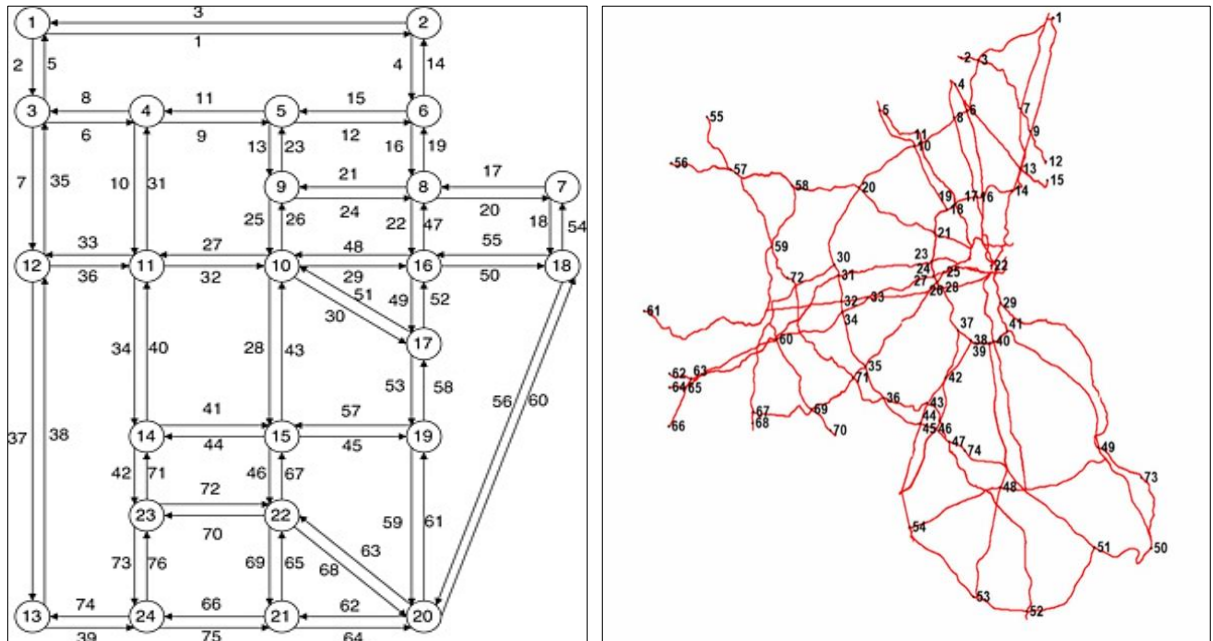

Fig. 3. Transportation Networks (a) Sioux Falls Network (Transportation Networks for Research Core Team, 2016) (b) East Massachusetts Network (US Census Bureau, 2015) (Huang and Kockelman, 2019)

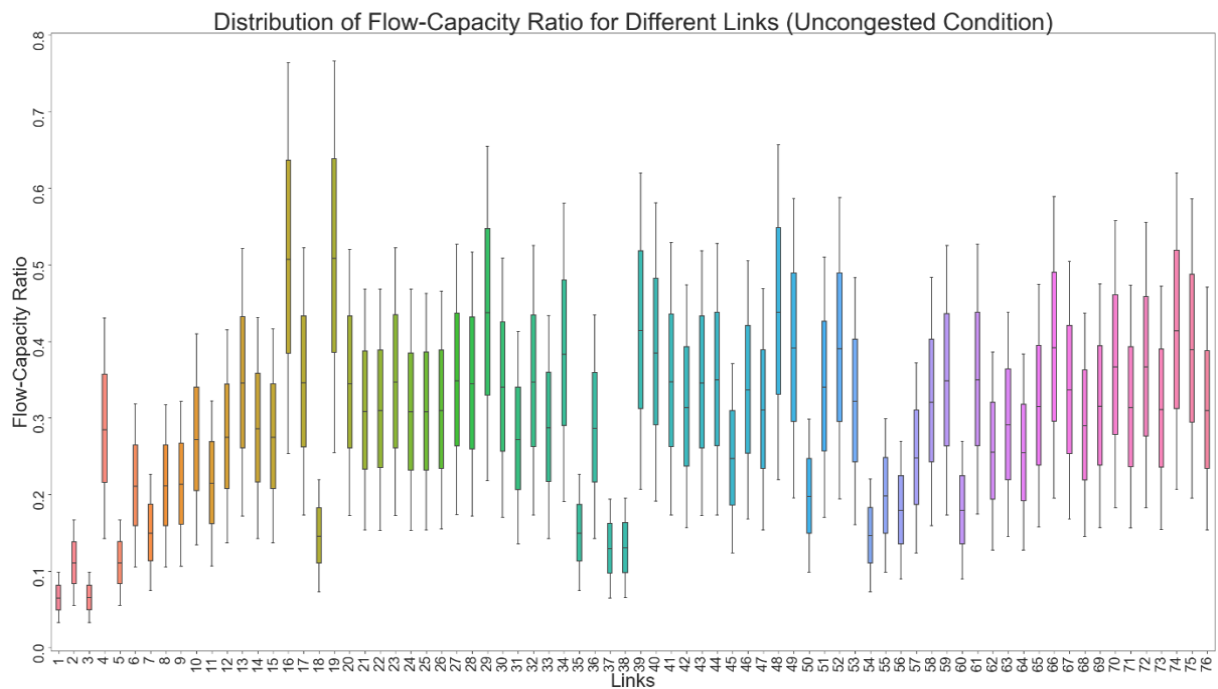

(a) Uncongested Condition

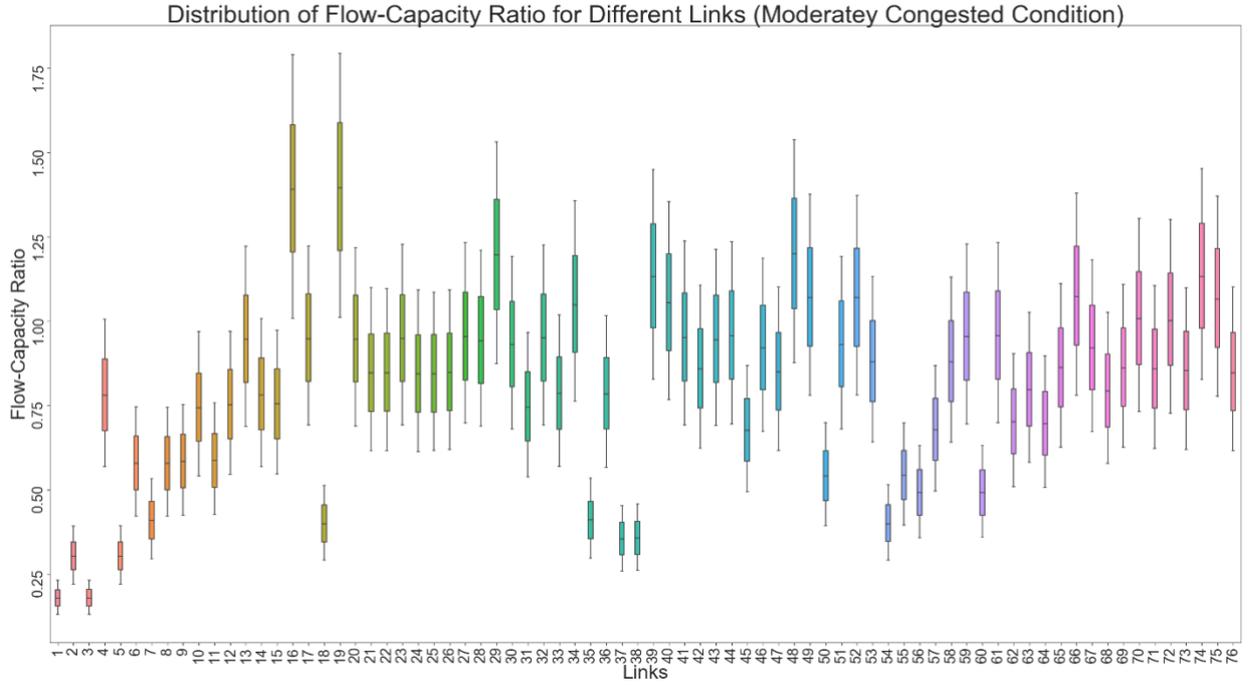

(b) Moderately Congested Condition

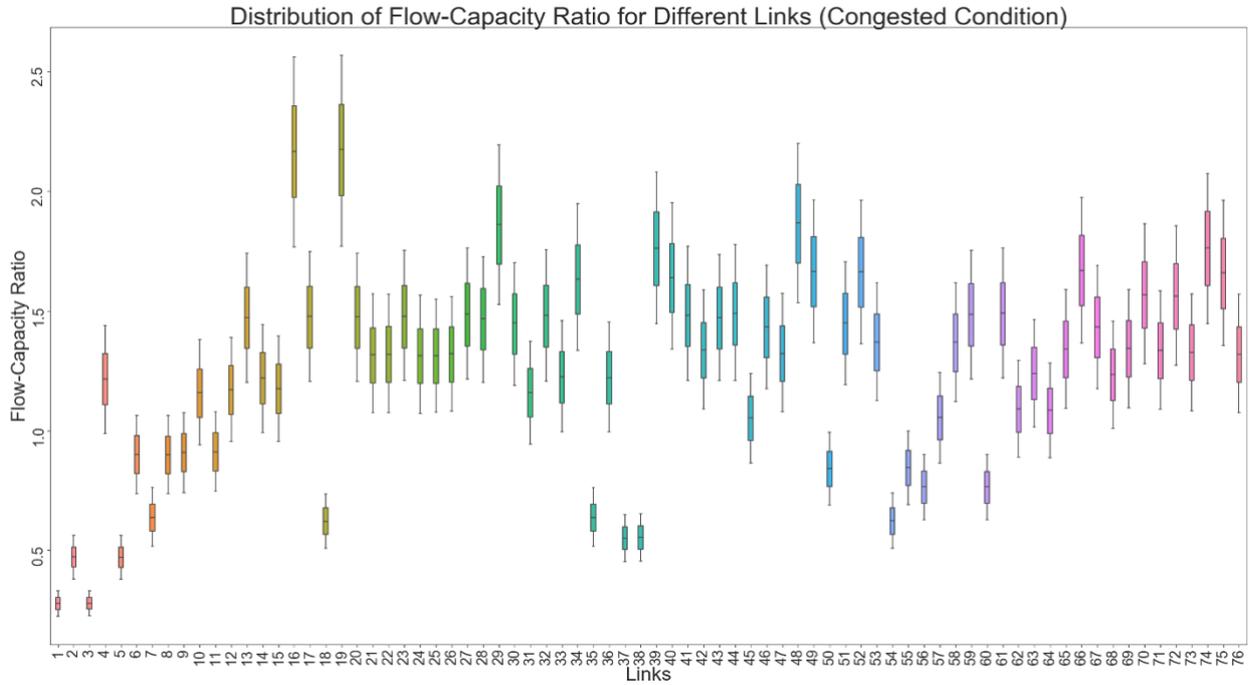

(c) Congested Condition

Fig. 4. Illustrates the distribution of flow-capacity ratio for different traffic conditions

## 4. RESULTS AND DISCUSSION

We implemented all the models using PyTorch ("PyTorch," 2016) library and train our model with dual NVIDIA Tesla V100 16GB PCIe GPU. Among the OD demand matrices, we use 70% (*n*=3,500) for training, 20% (*n*=1,000) for validation, and rest 10% (*n*=500) for testing the model. We train the model on training data and check the accuracy for the model on validation data set. Based on the validation accuracy, we tune the hyperparameters such as learning rate, types of activation functions (i.e., tanh, sigmoid etc.) and maximum number of iterations. We also check whether the model is overfitting or not. Once the final model is fixed, we test it on the test data set. We calculate Root Mean Squared Error (RMSE) and Mean Absolute Error (MAE) as performance measures to check the accuracy of the implemented model. Performance metrics are defined as,

$$RMSE = \sqrt{\frac{1}{m}\sum_{m=1}^{m}\frac{\sum_{\mathcal{E}=1}^{E}(F_m^{\mathcal{E}}-\hat{F}_m^{\mathcal{E}})^2}{E}} \quad (15)$$

$$MAE = \frac{1}{m}\sum_{m=1}^{m}\frac{\sum_{\mathcal{E}=1}^{E}|F_m^{\mathcal{E}}-\hat{F}_m^{\mathcal{E}}|}{E} \quad (16)$$

In Table 2, we report the performance of model on the test dataset. From the result, we find that both diffusion convolution and spectral convolution have similar accuracy, which is expected, since the spectral convolution operation is a special case of the diffusion convolution (Y. Li et al., 2018). Based on performance metrics values, we can conclude that the proposed approaches are performing well to capture the flow diffusion. RMSE and MAE values provides aggregated information (average over all the outputs) on the performance of the models, hence, we also estimated $R^2$ score. As shown in Table 2, for each model the $R^2$ score is nearly 1, indicating the accuracy of the model to learn traffic flows in the network. We have also compared between actual and estimated link flows for a given OD demand. Fig. 4 and Fig. 5 show that for both Sioux Falls and East Massachusetts networks the difference between actual and estimated link flow is quite low.

**Table 2** Model accuracy on the test datasets.

| Network | | Flow Propagation Function | Minimum Flow | Maximum Flow | Mean Flow | MAE | RMSE | % Error Over Mean Flow | $R^2$ Score |
|---|---|---|---|---|---|---|---|---|---|
| Uncongested Free Flow Condition | Sioux Falls | Random Walk | 642.8 | 7001.4 | 2447.4 | 8.6 | 11.9 | 0.350 | 0.9999 |
| | | Laplacian Graph | | | | 8.5 | 11.8 | 0.346 | 0.9999 |
| | | Spectral Graph | | | | 8.6 | 11.9 | 0.353 | 0.9999 |
| | East Massachusetts | Random Walk | 0 | 8220.1 | 1762.0 | 25.5 | 44.6 | 1.444 | 0.9991 |
| | | Laplacian Graph | | | | 25.2 | 44.5 | 1.431 | 0.9991 |
| | | Spectral Graph | | | | 25.5 | 44.6 | 1.449 | 0.9991 |
| Moderately Congested | Sioux Falls | Random Walk | 2562.9 | 16394.4 | 6704.5 | 23.2 | 31.7 | 0.345 | 0.9998 |
| | | Laplacian Graph | | | | 23.1 | 31.7 | 0.345 | 0.9998 |
| | | spectral Graph | | | | 23.4 | 32.1 | 0.350 | 0.9998 |
| | East Massachusetts | Random Walk | | | | 33.6 | 59.5 | 1.436 | 0.9991 |
| | | Laplacian Graph | 0 | 10858.7 | 2337.9 | 33.5 | 59.4 | 1.431 | 0.9991 |
| | | Spectral Graph | | | | 33.6 | 59.5 | 1.438 | 0.9991 |
| Congested Condition | Sioux Falls | Random Walk | | | | 36.0 | 48.4 | 0.346 | 0.9998 |
| | | Laplacian Graph | 4489.6 | 23437.3 | 10408.7 | 35.3 | 47.4 | 0.338 | 0.9998 |
| | | Spectral Graph | | | | 36.9 | 49.2 | 0.355 | 0.9999 |
| | East Massachusetts | Random Walk | | | | 38.8 | 68.9 | 1.438 | 0.9991 |
| | | Laplacian Graph | 0 | 12174.8 | 2698.7 | 38.7 | 68.8 | 1.439 | 0.9991 |
| | | Spectral Graph | | | | 39.3 | 69.1 | 1.4575 | 0.9991 |

From the results, we observe that: (i) a neural network can capture the traffic assignment of a network without any prior knowledge on user behavior; (ii) we can achieve a better accuracy with an appropriate representation of the physical process of flow propagation.

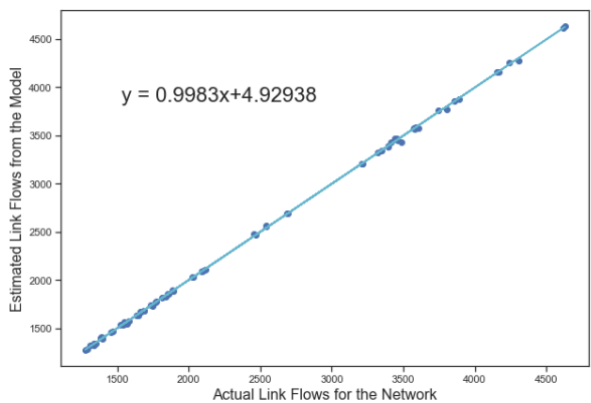
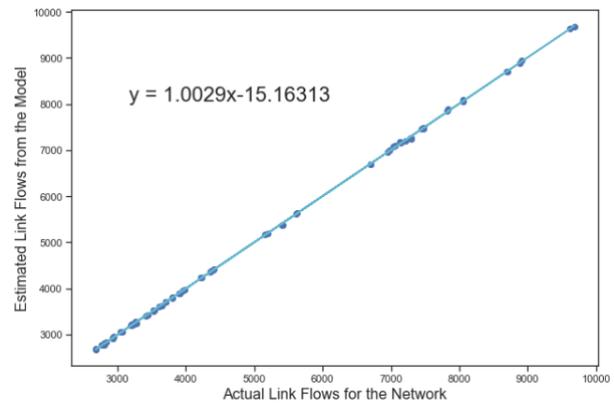

(a) Uncongested Condition    (b) Moderately Congested Condition

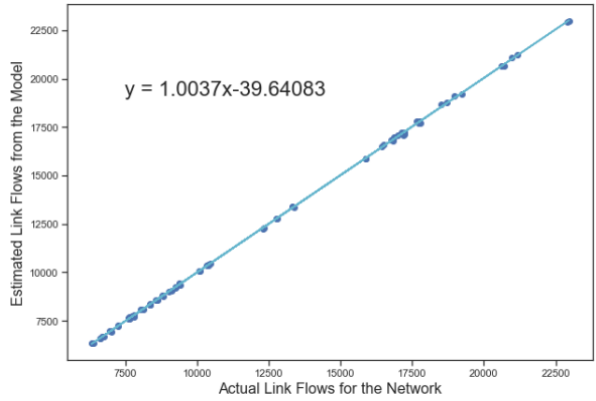

(c) Congested Condition

Fig. 4. Comparison between actual and estimated link flows for a given OD demand Sioux Falls Network

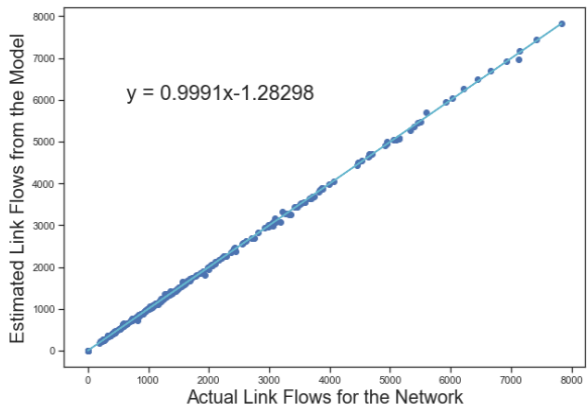

(a) Uncongested Condition

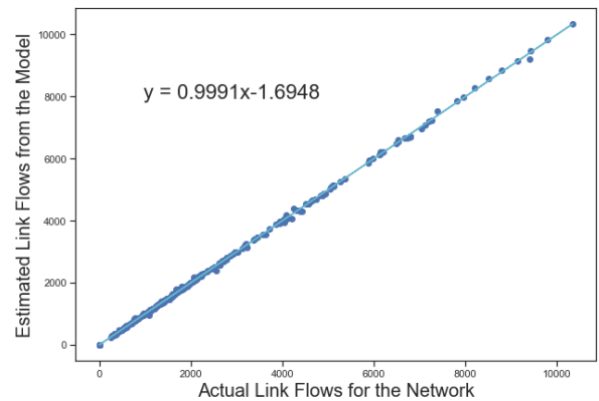

(b) Moderately Congested Condition

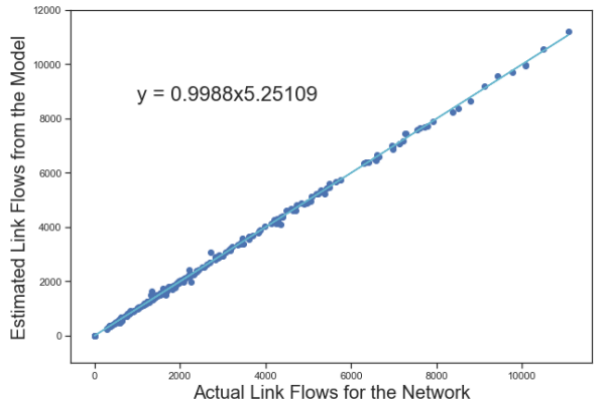

(c) Congested Condition

Fig. 5. Comparison between actual and estimated link flows for a given OD demand East Massachusetts network

## 4.1 Result interpretation

To understand how well the model has learned the flow propagation for the given networks, we perform network topology analysis for the Sioux Falls network based on the betweenness centrality of the nodes. Betweenness centrality of a node indicates the fraction of the total number of shortest paths passing through that node (Brandes, 2001), which means a node with a higher value of betweenness centrality will have a higher number of shortest paths incidence on that particular node.

For a given OD demand matrix, to estimate betweenness centrality values we need to find the shortest paths for all pairs of origin-destination nodes at user equilibrium. However, the shortest path for a pair of origin and destination nodes depends on link travel times. In our analytical solutions (i.e., using Frank-Wolfe algorithm), we use BPR travel time function (see equation 24) to update link travel times. Based on the updated travel time, we find the shortest paths for assigning traffic to the network. These steps continue iteratively prior to reaching an user equilibrium solution, when travel time for all the used paths remain same for a given O-D pair. Hence, we use the same function to estimate user equilibrium travel time ($t_{i,j}$) for all the links of the network.

$$t_{i,j} = t_{i,j}^0 \left(1 + 0.15 \left(\frac{f_{i,j}}{c_{i,j}}\right)^4\right) \quad (17)$$

where $\frac{f_{i,j}}{c_{i,j}}$ indicates the flow-capacity ratio for a given link. Based on the estimated links' travel time from equilibrium link flows we find the shortest paths and estimate the betweenness centrality for all the nodes. We apply this approach on all the training OD demand samples. Fig. 6 shows the distribution of betweenness centrality values for each node of the Sioux Falls Network across different traffic conditions. From the figures we find that except nodes 1, 2, 9, and 23, all the nodes have higher betweenness centrality values. Nodes 6, 8, 12, 15, 16, and 18 are the most critical for the Sioux Falls network. Moreover, we find that for congested condition (i.e., higher demand), variations of betweenness centrality values are higher compared to moderately and uncongested conditions.

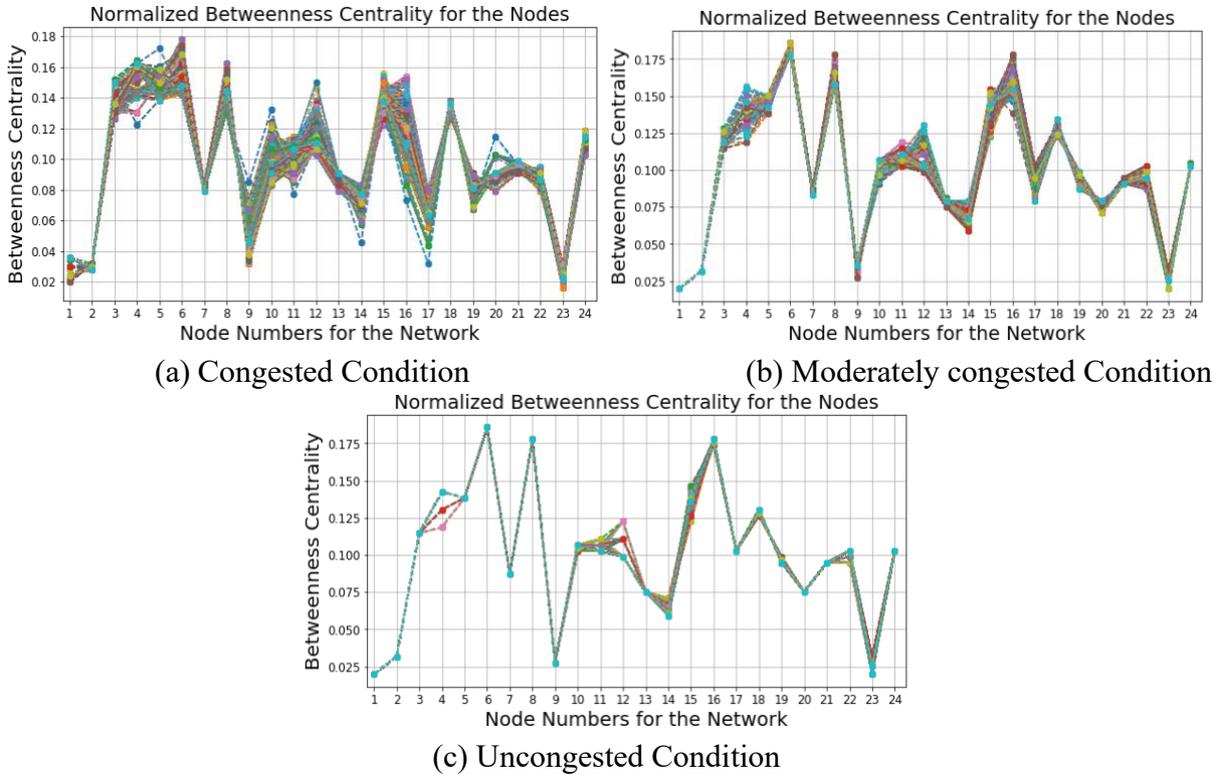

Fig. 6. Variations of Betweenness Centrality for different nodes at different traffic condition (Sioux Falls network)

From the BPR function, we find that, if flow capacity ratio is less than 1.0, there will be no significant change in travel time with link flow variations or demand variations. However, when the flow capacity ratio is greater than 1.0, travel time will vary significantly with demand variations (i.e., link flow variations). Consequently, the shortest paths will change abruptly (i.e., shortest path are nor stable) leading to significant variations in betweenness centrality for all the nodes. In our case, for congested condition, the flow capacity ratio mostly varies from 1.0 to 2.0, thereby we observe a significant variation of betweenness centrality for different nodes. Whereas for moderately congested and uncongested conditions, since the flow capacity ratio is mostly less than 1.0 so we, do not see significant variations in Betweenness Centrality.

In the proposed model, we assign parameters $(W_q)$ to learn the flow propagation from nodes into adjacent neighboring links. We assume that the weight parameters associated with critical nodes will be higher and will vary significantly due to the changes in betweenness centrality of nodes. In other words, the weight parameter associated with a node is likely be correlated with the betweenness centrality value of the node.

In Fig. 7, we plot the weight distributions for node-link flow propagation inside Sioux Falls network at different traffic conditions. Since for uncongested and moderately congested conditions, the variations of betweenness centrality for different nodes are similar, the distributions of weight parameter $(W_q)$ are also similar. In both cases, the critical nodes 6, 8, 15, and 16 have high positive weights (Fig. 7 (a) & (b)). For congested condition, the variation of betweenness centrality over different nodes are higher, thereby the weight parameters vary significantly compared to uncongested and moderately conditions (Fig. 7(c)). In a congested condition, the model cannot identify the critical nodes from all nodes to pass the traffic efficiently. This could be a possible reason that the model does not give higher positive weights for critical nodes similar to uncongested and moderately congested conditions.

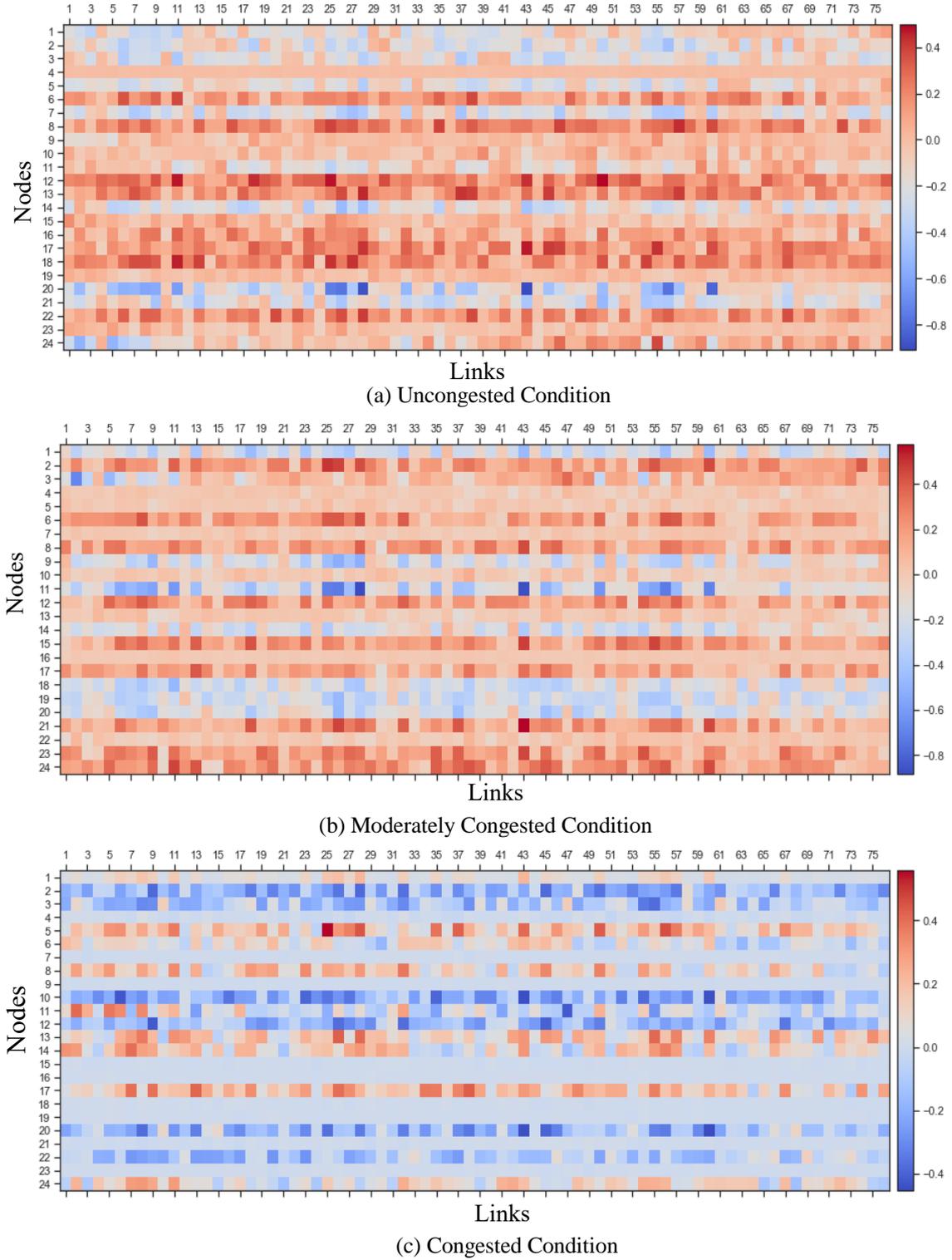

Fig. 7. Distribution of weight ($w_q$) over different traffic conditions (Sioux Falls network)

## 4.2 Stability of the Solution

We train the model using mean squared error as the loss function. At each iteration, the model estimates the mean squared error for the estimated flows ($\hat{F}_m^\varepsilon$) and the actual flows ($F_m^\varepsilon$) of the network. Afterward, the gradient of the loss function is backpropagated to adjust the weights to reduce loss function value. The loss function can be defined as,

$$L_m = Loss\left(F_m^{\mathcal{E}}, \hat{F}_m^{\mathcal{E}}\right) \qquad (18)$$

$$MSE = \frac{1}{m}\sum_{m=1}^{m}\frac{\sum_{\mathcal{E}=1}^{E}(F_m^{\mathcal{E}} - \hat{F}_m^{\mathcal{E}})^2}{E} \qquad (19)$$

where, $Loss(.)$ is the function to estimate the error between the actual ($F_m^{\mathcal{E}}$) and estimated values ($\hat{F}_m^{\mathcal{E}}$) and $\mathcal{E}$ denotes the set of links for the network. In this study, we estimate mean square error (MSE) as a loss function.

To check the stability of solution, we observed the training and test loss values for the model (Fig. 8). We train each model for 10,000 iterations to check variation of train and validation loss values. We find that it takes about 2000 iterations for the model to converge to a stable solution, after that there are merely any variations in loss values. Moreover, after 10,000 iterations the loss function value for the validation data gradually start increasing. From 2,000 to 10,000 iterations the difference in MSE values varies from 4 to 9, indicating a stable solution with minimal variance. We experimented with two different optimizers to train the model: root mean square propagation (RMSProp) and adaptive moment estimation (ADAM) optimizer. Among these two, RMSProp takes less iterations (~1500 iterations) to converge (i.e. similar train and validation error) compared to ADAM optimizer (~2000 iterations). However, ADAM optimizer gives more stable solutions, which means MSE values for train and validation data almost remain same after convergence (i.e., after 2000 iterations). Whereas, for RMSProp optimizer we observe slight variations in MSE values for both train and validation data samples even after convergence (i.e., after 1500 iterations).

We also check the computation time required to train the models. It takes about 19 minutes to train the models on Sioux Falls network for 10,000 iterations, while for East Massachusetts network it takes 30 minutes. So, our approach performs reasonably well to estimate network level traffic flows with less computation time.

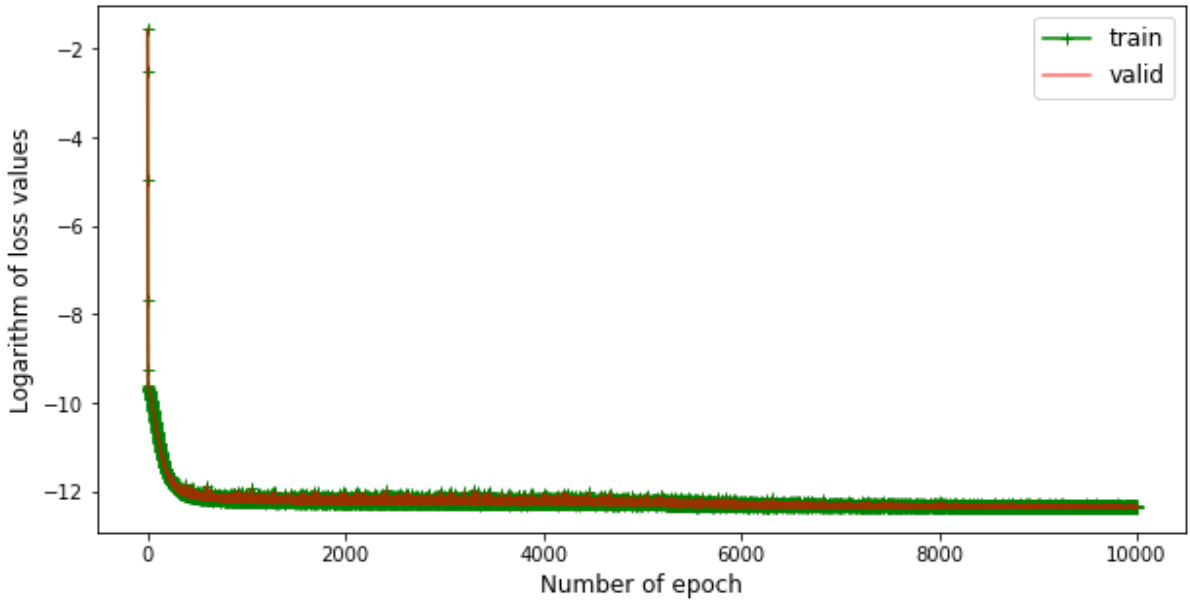

(a) Congested Condition

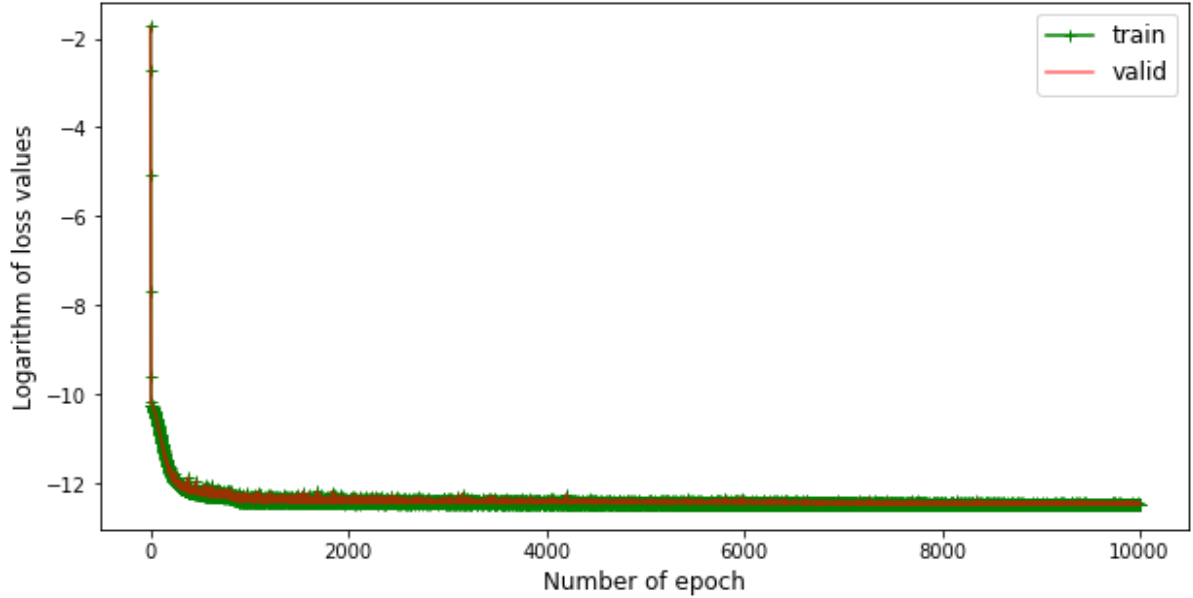

(b) Moderately Congested Condition

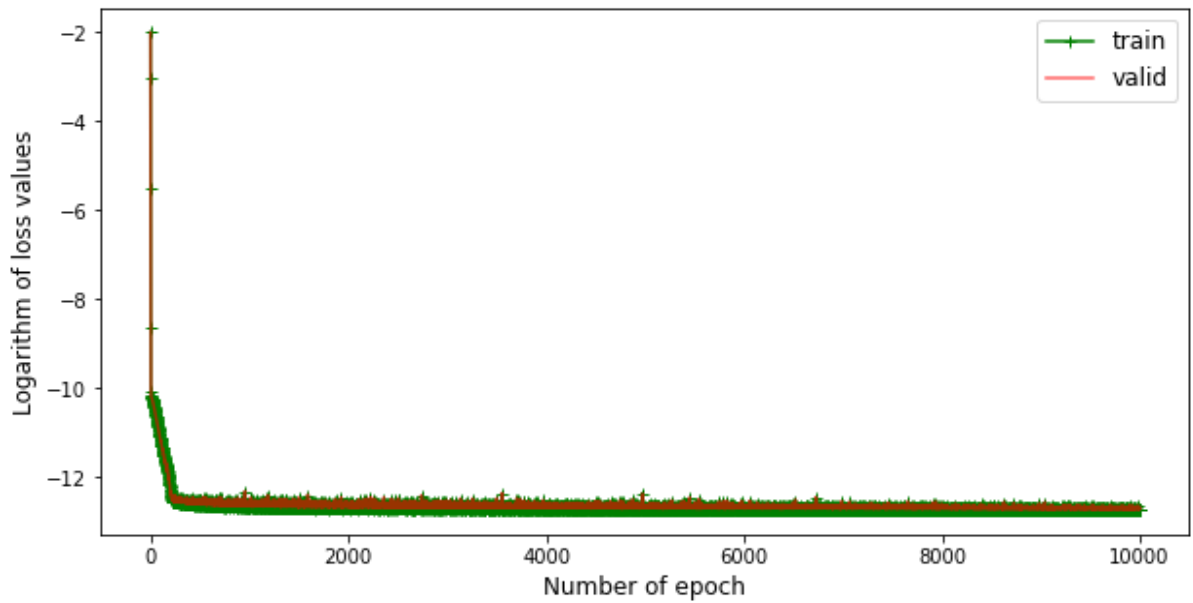

(c) Uncongested Condition

Fig. 8. Loss Function Values for Laplacian Spectral Graph (Sioux Falls network)

## 5. CONCLUSIONS

In this study, we present a data-driven formulation of traffic assignment problem based on learning traffic flow patterns of a transportation network from origin destination (OD) travel demand variations. Adapting graph convolution approach, we develop a deep learning architecture to solve this traffic assignment problem by capturing the diffusion of OD demand inside the network. To efficiently represent the diffusion process of multiple OD demands from nodes to neighboring links, we customize the traditional graph convolutional neural network and introduce the concept of learning-based assignment (i.e., routing matrix) of OD demands to generate link flows. Finally, we provide experimental evidence on the validity of the approach by training the model to learn the user equilibrium traffic flows for Sioux Falls and East Massachusetts networks. The experiment results show that the implemented GCNN model can capture the user equilibrium traffic flow of the network very well with less than 2% mean absolute difference between the actual and estimated link flows under varying congested conditions. Moreover, when the training of the model is complete, it can instantaneously determine the traffic flows of a large-scale network. Hence this approach can overcome the challenges of deploying traffic assignment models over large-scale networks. Furthermore, this method is completely data-driven without requiring any assumption on user behavior. Thus, it will improve the reliability and stability of traffic assignment solutions.

To extend this framework towards a data-driven dynamic traffic assignment method, we need to consider the representation of the physical process of flow propagation to account travel time variations. In addition, existing data sources are not widely available to researchers to infer travel demand at a higher spatiotemporal resolution. Alternative data augmentation approach can be explored to prepare the travel demand data. Future research should consider these issues to develop a framework that can solve the dynamic traffic assignment problem and prepare demand data that can be fed into such frameworks.

A data-driven network modeling approach is warranted in the era of big data. Ubiquitous use of mobile phones, availability of GPS based vehicle trajectory data, emerging connected and automated vehicle data, and so on will give us the opportunities to model travel demand and traffic flows at a high spatio-temporal resolution. The proposed deep learning architecture for solving a traffic assignment learning problem is an initial step towards exploiting such high-fidelity data for data-driven network modeling research.

**Appendix A. Modeling Traffic Flows using Spectral Graph Convolution**

In spectral graph convolution, a spectral convolutional filter is used learn traffic flow patterns inside a transportation network in repose to travel demand variations. The spectral filter is derived from spectrum of the Laplacian matrix, which consist of eigenvalues of the Laplacian matrix. So, to construct the spectrum, we must calculate the eigenvalues of` the Laplacian matrix. For a symmetric graph, we can compute the eigenvalues using eigen decomposition of the Laplacian matrix. In this problem, we consider the transportation network as a symmetric directed graph; same number of links getting out and getting inside a node, which means the in degree and out degree matrix of the graph is similar. Thus, the Laplacian matrix of this graph is diagonalizable as follows using eigen decomposition,

$$\boldsymbol{L_w} = \boldsymbol{U \Lambda U^T} \tag{20}$$

where, $\boldsymbol{\Lambda}$ is a diagonal matrix with eigenvalues, $\lambda_0, \lambda_1, \lambda_2, \ldots, \lambda_N$ and $\boldsymbol{U}$ indicates the eigen vectors, $u_0, u_1, u_2, \ldots, u_N$. Eigen values represent characteristics of transportation network in terms of strength of a particular node based on its position, distance between adjacent nodes, dimension of the network. The spectral graph convolution filter can be defined as follows,

$$g_\theta(\Lambda) = \sum_{k=0}^{K-1} \theta_k \Lambda^k \tag{21}$$

where, $\theta$ is the parameter for the convolution filter shared by all the nodes of the network and $K$ is the size of the convolution filter. Now the spectral graph convolution over the graph signal ($X$) is defined as follows,

$$g_\theta * X = g_\theta(L_w)X = g_\theta(U\Lambda U^T)X = Ug_\theta(\Lambda)U^T X = \sum_{k=0}^{K-1} \theta_k U\Lambda^k U^T X$$

$$g_\theta * X = \sum_{k=0}^{K-1} \theta_k L_w^k X \tag{22}$$

According to spectral graph theory, the shortest path distance i.e. minimum number of links connecting nodes $i$ and $j$ is longer than $K$, such that $L^K(i,j) = 0$ (Hammond et al., 2011). Consequently, for a given pair of origin ($i$) and destination ($j$) nodes, a spectral graph filter of size $K$ has access to all the nodes on the shortest path of the graph. It means that the spectral graph convolution filter of size $K$ captures flow propagation through each node on the shortest path. So, the spectral graph convolution operation can model the interdependency between a link and its $i$ th order adjacent nodes on the shortest paths, given that $0 \leq i \leq K$.

The computational complexity of calculating $L_w^k$ is high due to K times multiplication of $L_w$. A way to overcome this challenge is to approximate the spectral filter $g_\theta$ with Chebyshev polynomials up to $(K-1)$th order (Hammond et al., 2011). Defferrard et al. (Defferrard et al., 2016) applied this approach to build a $K$-localized ChebNet, where the convolution is defined as,

$$g_\theta * X \approx \sum_{k=0}^{K-1} \theta_k T_k(\bar{L})X \tag{23}$$

in which, $\bar{L} = 2L_{sym}/\lambda_{max} - I$. $\bar{L}$ represents a scaling of graph Laplacian that maps the eigenvalues from $[0, \lambda_{max}]$ to $[-1,1]$. $L_{sym}$ is defined as symmetric normalization of the Laplacian matrix $D_w^{-1/2} L_w D_w^{-1/2}$. $T_k$ and $\theta$ denote the Chebyshev polynomials and Chebyshev coefficients. The Chebyshev polynomials are defined recursively by $T_k(\bar{L}) = 2xT_{k-1}(\bar{L}) - T_{k-2}(\bar{L})$ with $T_0(\bar{L}) = 1$ and $T_1(\bar{L}) = \bar{L}$. These are the basis of Chebyshev polynomials. Kipf and Welling (Kipf and Welling, 2016) simplified this model by approximating the largest eigenvalue $\lambda_{max}$ of $\bar{L}$ as 2. In this way, the convolution becomes,

$$g_\theta * X = \theta_0 X - \theta_1 D_w^{-1/2} A_w D_w^{-1/2} X \tag{24}$$

where, Chebyshev coefficient, $\theta = \theta_0 = -\theta_1$, All the detail about the assumptions and their implications of Chebyshev polynomial can be found in (Hammond et al., 2011). Now the simplified graph convolution can be written as follows,

$$g_\theta * X = \theta(I + D_w^{-1/2} A_w D_w^{-1/2})X \tag{25}$$

Since $I + D_w^{-1/2} A_w D_w^{-1/2}$ has eigenvalues in the range [0, 2], it may lead to exploding or vanishing gradients when used in a deep neural network model. To alleviate this problem, Kipf et al. (Kipf and Welling, 2016) use a renormalization trick by replacing the term $I + D_w^{-1/2} A_w D_w^{-1/2}$ with $\bar{D}_w^{-1/2} \bar{A}_w \bar{D}_w^{-1/2}$, with $\bar{A}_w = A_w + I$, similar to adding a self-loop. Now, we can simplify the spectral graph convolution as follows,

$$g_\theta * X = \Theta \left( \bar{D}_w^{-1/2} \bar{A}_w \bar{D}_w^{-1/2} \right) X \tag{26}$$

here, $\Theta \in R^{N \times N}$ indicates the parameters of the convolution filter to be learnt during training process. From Equation 21, we can observe that spectral graph convolution is a special case of diffusion convolution (Y. Li et al., 2018), only difference is that in spectral convolution we symmetrically normalized the adjacency matrix.

## ACKNOWLEDGMENT

This study was supported by the U.S. National Science Foundation through the grant CMMI #1917019. However, the authors are solely responsible for the facts and accuracy of the information presented in the paper.